\documentclass[10pt,journal,compsoc]{IEEEtran}

\usepackage[nocompress]{cite}

\usepackage[pdftex]{graphicx}
\usepackage[normalem]{ulem} 
\usepackage{booktabs} 
\usepackage[usenames, dvipsnames]{color} 
\usepackage{footnote}

\begin{document}

\title{SpaRTA \\
Tracking across occlusions via global partitioning of 3D clouds of points}

\author{ Andrea Cavagna, Stefania Melillo, Leonardo Parisi, Federico Ricci-Tersenghi
        
    \IEEEcompsocitemizethanks{
        \IEEEcompsocthanksitem A.~Cavagna, S.~Melillo and L.~Parisi are with CoBBS Lab (Collective Behaviour in Biological Systems, http://www.cobbs.it) at CNR--ISC (National Research Council - Institute for Complex Systems) UOS Sapienza, Rome, Italy. 
        \IEEEcompsocthanksitem L.~Parisi is currently with DIMA (Department of Mechanical and Aerospace Engineering) at Sapienza University of Rome, Italy.
        \IEEEcompsocthanksitem F.~Ricci-Tersenghi is with Physics Department at Sapienza University of Rome, Italy 
    }
    
}

\IEEEtitleabstractindextext{

\begin{abstract}
   Any $3D$ tracking algorithm has to deal with occlusions: multiple targets get so close to each other that the loss of their identities becomes likely. In the best case scenario, trajectories are interrupted, thus curbing the completeness of the data-set; in the worse case scenario, identity switches arise, potentially affecting in severe ways the very quality of the data. Here, we present a novel tracking method that addresses the problem of occlusions within large groups of featureless objects by means of three steps: i) it represents each target as a cloud of points in $3D$; ii) once a $3D$ cluster corresponding to an occlusion occurs, it defines a partitioning problem by introducing a cost function that uses both attractive and repulsive spatio-temporal proximity links; iii) it minimizes the cost function through a semi-definite optimization technique specifically designed to cope with link frustration. The algorithm is independent of the specific experimental method used to collect the data. By performing tests on public data-sets, we show that the new algorithm produces a significant improvement over the state-of-the-art tracking methods, both by reducing  the number of identity switches and by increasing the accuracy of the actual positions of the targets in real space.
\end{abstract}

}

\maketitle

\IEEEdisplaynontitleabstractindextext
\IEEEpeerreviewmaketitle

\section{Introduction}

\IEEEPARstart{T}{racking} large groups of targets in $3D$ space is a challenging topic, which is particularly relevant in the field of turbulence \cite{ouellette2006quantitative}, collective animal behavior \cite{wu2016global}, \cite{cheng2015novel} and social sciences \cite{moussaid2012traffic}, \cite{wen2016multi} as well as in robotics \cite{michel2007gpu} and autonomous mobility \cite{ess2010object}. The  technological progress of the last decades gave a boost to the development of new experimental strategies to collect $3D$ data, such as RGB--D, multicamera, lidar and radar systems. Nowadays the effort of a part of the computer vision community is directed towards finding general high-performance tracking methods. 

The crucial point of all tracking algorithms is how to handle occlusions that arise every time that two or more objects get too close in $3D$ space to be detected as multiple targets. This kind of ambiguities are particularly severe when dealing with featureless objects (objects that cannot be identified by any feature such as shape or color) and with large and dense groups of targets, where the chance to get in $3D$ proximity is high. Occlusions hinder in a twofold way the quality of the retrieved trajectories: loss of one or more of the targets involved into the occlusions and a potential switch of identities. While the first drawback generally affects at the quantitative level the completeness of our tracking effort, the second one may severely change at the qualitative level the results of any data analysis.

In this paper we propose a novel tracking method called SpaRTA (Spatiotemporal Reconstruction Tracking Algorithm), which is able to solve $3D$ occlusions identifying each target during the occlusions and producing negligible switches of identities. SpaRTA is meant to work on objects detected as $3D$ clouds of points, regardless of the system used to collect the data, such as RGB--D, multicamera, lidar or radar. The core ideas of the methods are the following: i) SpaRTA reconstructs the $(3D+1)$ spatio-temporal volume (where $3D$ is the spatial dimension and $(+1)$ represents the time dimension) occupied by each target during the entire acquisition as a cloud of points; ii) when an occlusion arises, SpaRTA tackles the problem of splitting it into different objects by defining a partitioning problem that uses both attractive and repulsive links depending on the distance in space and time among the points belonging to the occlusion; iii) as the superposition of attractive and repulsive links gives rise to frustration, namely to the emergence of many local minima of the partitioning cost function, SpaRTA uses an optimization method inspired on Semi-Definite Programming (SDP) techniques developed in the context of statistical physics of disordered systems \cite{ricci2016performance}, to find the optimal partition (i.e. ground state of the cost function), thus finally splitting the occlusion into the actual different targets composing it.

\medskip

SpaRTA was tested on data of large groups of animals collected in the field with a multicamera system. This kind of data are a good benchmark for $3D$ tracking methods because they are particularly hard to track: they are characterized by frequent occlusions, lasting several frames, and by a low spatial resolution such that targets appear as objects without any recognizable feature. The only limitation of these data is that the production of ground truth trajectories to evaluate the tracking result is quite difficult and time--consuming, and it is then hard to give a quantitative evaluation of the quality of the resulting set of trajectories. This is the reason why there are very few public data-sets to be used as benchmark. To the best of our knowledge, the only available two public data-sets of featureless objects collected via a multicamera system are published in \cite{wu2014thermal}. We tested SpaRTA on these data-sets showing the high performance of the proposed algorithm in terms of the quality of the retrieved trajectories.

\section{Related works}\label{sec:relatedWorks}

Since the seminal work of Reid \cite{reid1979algorithm}, several $3D$ tracking strategies have been proposed in the past forty years. However, despite this strong effort, only few methods are designed to track large and dense systems of featureless objects and the research on this topic is still very much ongoing, especially for what concerns the solution of occlusions. 
There are two fundamentally different ways to represent the targets an algorithm wants to track: on one hand, we can associate to each target at a certain instant of time one single identity and spatial position (typically the baricenter) -- we will call this case Single Point (SIP) representation; on the other hand, we can associate to a target a dense Cloud of Points (COP), representing the the full spatial volume of the target at that instant of time, see Appendix A.

The SIP representation is typically adopted within the context of multicamera data-taking systems, in which sets of $2D$ single objects positions have to be turned into $3D$ positions and trajectories. 
One way to achieve this is to first track the objects in each camera and then match the $2D$ trajectories  across cameras to retrieve the corresponding $3D$ trajectories  (the so-called Tracking-Reconstruction (TR) route). Working in the $2D$ space of the cameras, TR algorithms have to deal not only with $3D$ occlusions but also with $2D$ occlusions, namely ambiguities due to targets getting in spatial proximity only on the image plane but not in the $3D$ space. $2D$ occlusions produce bifurcations of the $2D$ trajectories, and hence a high intrinsic complexity due to the proliferation of the $2D$ trajectories to be matched across the cameras. Several different strategies to prune the set of $2D$ trajectories and reduce the complexity of this approach have been devised \cite{attanasi2015GReTA}, \cite{cox1996efficient}, \cite{wu2011efficient}, \cite{wu2011automated}, \cite{wu2016global} and \cite{liu2012automatic}.
Conversely, yet still within the SIP representation, one can first reconstruct single objects turning them into $3D$ objects (by matching their identities across the cameras), and then track them in $3D$ space (the so-called Reconstruction-Tracking (RT) route). Working directly in the $3D$ space, RT methods are not affected by $2D$ occlusions that are naturally solved when matching objects across the cameras (see Appendix B), hence their complexity is naturally quite lower than the TR methods; however, RT strategies are typically more prone to creating false $3D$ objects. The SIP-RT approach has been explored in small groups of objects \cite{tyagi2007fusion} \cite{li2002relaxation} \cite{dockstader2001multiple} \cite{cheng2015novel}, and it is not clear how these approaches perform in case of dense groups. The most advanced SIP-RT method has been proposed in \cite{wu2009tracking}, which can successfully track large groups.

The SIP representation has some drawbacks, especially in dense systems, where occlusions are frequent. More specifically, whenever two or more objects are part of a single occlusion, the SIP method associates one single position to all of them, causing: i) loss of the actual targets individual identities; ii) potential identity switches after the occlusion; iii) an inaccurate positioning of the targets in $3D$ space (see Appendix A).

The COP representation, on the other hand, allows to associate to each object (at each instant of time) a dense cloud of $3D$ points and not only its baricenter. This dense representation reconstructs the actual volumes occupied by the detected objects. Because COP discards no information about the actual targets volumes, and therefore creates no simplified identities, it is more suited to prevents identity switches (problems i) and ii) above) and definitely more accurate to locate the $3D$ positions of the targets.
Besides, COP-based tracking (unlike SIP) is not forcibly embedded within a multicamera framework and it is therefore a significantly more general approach than SIP: indeed, COP tracking has been used in RGB--D systems  \cite{RGBD-Overview2013}, lidar \cite{Lidar-Asvadi2016} and radar \cite{RADAR-Mobus2003}. In the COP context, occlusions have been tackled by different techniques, \cite{RGBD-2012}, \cite{ZHANG2013126}, \cite{Lidar-Choi2013}, \cite{RADAR-2004}, which, however are designed for the specific nature of the data to be tracked. Here, we will introduce a novel COP-based tracking method designed to be as general as possible.

\section{Overview of the method}\label{sec:overview}

SpaRTA works with $(3D+1)$ (space + time) clouds of points representing the $3D$ volume occupied in time by a group of moving objects, without any limitation on the $3D$ system used to collect the data\footnote{Note that when using a multicamera acquisition method, data are not directly obtained as $3D$ clouds of points (unlike data acquired with RGB-D, lidar or radar systems); hence, in that specific case, targets images are converted into $3D$ clouds of points through a pre--processing procedure as the one described in \cite{cavagna2016towards} and in Appendix C.}. The goal of the algorithm is to partition the $(3D+1)$ cloud of points in $(3D+1)$ sub--clouds, each corresponding to the trajectory of a single target. 

\begin{figure}[h!]
\begin{center}
\includegraphics[width=1.0\linewidth]{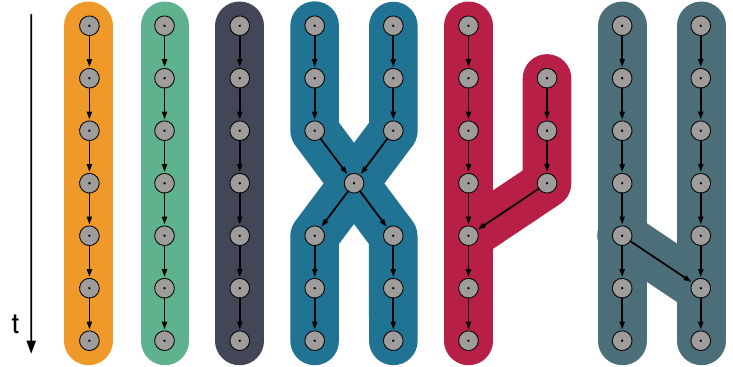}
\end{center}
\caption{\textbf{Clusters graph.} Grey circles represent the clouds of points that are then dynamically linked. The connected components are highlighted in different colors. The first three components from the left represent single trajectories, since they are made of only one--to--one linked clusters. The last three components are, instead, ambiguous because they have at least one node with more than one link from the past or to the future.}
\label{fig:clusters_graph}
\end{figure}

SpaRTA can be broken into the following steps:

\medskip

\textbf{1~-- Building the graph.} The cloud of $3D$ points is first clustered in space at a static level (fixed instant of time): a clustering algorithm based on the $3D$ nearest neighbor distance \cite{Rokach2005} is used to detect the well--separated dense cloud of points (clusters), which may represent the detected objects at each instant of time. These reconstructed $3D$ clusters are then connected in time through a dynamical linking procedure, see Section~\ref{sec:graph_construction} and Appendix~D for further details. In this way, we create a set of $(3D+1)$ clouds of points representing the volumes occupied by the objects during the event, actually building the graph shown in Fig.~\ref{fig:clusters_graph} with $3D$ clusters as nodes and links as edges. 

\medskip

\textbf{2~-- Tackling the occlusions.} A breadth--first search routine \cite{hopcroft1973algorithm} is used to identify the connected components of the clusters graph, which should represent the trajectories of the detected targets. In the ideal situation where $3D$ occlusions do not occur, each connected component is made of one single target at each instant of time, therefore it is made only of one--to--one linked nodes, see Fig.~\ref{fig:clusters_graph}. However, in the more realistic situation where $3D$ occlusions do occur, two or more objects may belong to the same connected component, sharing one or more nodes, as in the last three cases in Fig.~\ref{fig:clusters_graph}. These connected components, with at least one multi-link, are due to occlusions, and they must be solved. The philosophy of SpaRTA is to break up the ambiguous connected components into different partitions, each corresponding to the trajectory of a single actual target, by defining and solving an optimization problem. First, a graph is built, whose nodes are the points belonging to the occlusion cluster and whose links depend on the distance/proximity in space and time of these points. The crucial idea, here, is to use both {\it attractive} (positive) links connecting points that are close enough to suggest high probability of belonging to the same actual target, but at the same time to penalize with {\it repulsive} (negative) links pairs of points that are too far from each other, compared with intrinsic space-time scales of the data. Once the graph is built, SpaRTA defines a cost function given by the negative sum of all links in each candidate partition, in such a way that the global minimum of this function corresponds to the optimal partitioning of the occlusion into {\it bona fide} $3D$ targets. The presence of both attractive and repulsive links is crucially functional in associating the correct partition to the actual targets; on the other hand, using competing links is known to increase steeply the complexity of an optimization method, by creating a proliferation of sub-optimal solutions (local minima) \cite{mezard2009information}. To deal with this problem, SpaRTA uses an optimization routine inspired on Semi-Definite Programming (SDP) techniques, that are known to work efficiently in disordered systems whose complexity is severe \cite{ricci2016performance},  \cite{javanmard2016phase}.  See Section \ref{sec:occlusions_solution} for details.

\medskip

\textbf{3~-- Identifying \textit{3D} trajectories}. Each non--ambiguous connected component identified in Step 2 above, represents the $3D$ volume occupied by a single target during the dynamics of the system. However, for many practical purposes it is not convenient to work with $3D$ volumes, which may be hard to be handled, and it is more desirable to associate a single $3D$ position to each object at each instant of time. Thus, we associate to each cluster its $3D$ baricenter position, i.e. average $3D$ coordinates of the cluster points, and we define the trajectories as the time sequence of the baricenter coordinates. Notice that this is {\it not} the same as defining objects positions through their baricenter in the SIP framework, because here all occlusions have been already solved, hence the baricenter is indeed a fair tool to locate the target position; on the contrary, baricenters fail in SIP whenever one object corresponds to several targets into an occlusion.

\section{Method details}

In this section we describe in detail how we handle the $(3D+1)$ cloud of points to build the graphs and how we solve $3D$ occlusions.

\subsection{Building the graph}\label{sec:graph_construction}

The $(3D+1)$ cloud of points is first analyzed at a static level to identify well--separated clusters, which may represent single objects or multi-objects during an occlusion. To this aim, we use a standard clustering algorithm based on the $3D$ nearest neighbor distance \cite{Rokach2005}: two reconstructed $3D$ points, $Q_1$ and $Q_2$ belong to the same cluster, $C$, if their $3D$ distance $d(Q_1,Q_2)$ is smaller than $r_1$, with $r_1$ equal to the median of the targets nearest neighbor distance. The complexity of brute force implementation of this procedure is $O(M^2)$, with $M$ being the average number of $3D$ points at each frame, but we lower it to $O(M)$ by using the space--partitioning technique of \cite{turk1989interactive}.

\begin{figure}[h!]
\begin{center}
\includegraphics[width=1.0\linewidth]{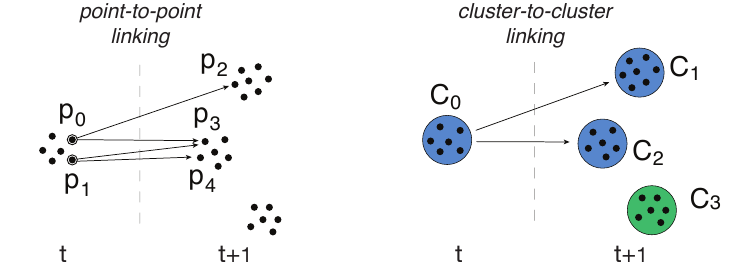}
\end{center}
\caption{\textbf{Cluster graph construction: dynamic linking.} At a generic frame $t$, a point--to--point multi--linking procedure is performed: $p_0$ at time $t$ is connected to $p_2$ and $p_3$ at frame $(t+1)$, while $p_1$ at frame $t$ is connected with $p_3$ and $p_4$. These point--to--point links are then used to define cluster--to--cluster links: two clusters are connected if there exists at least one point--to--point link between points belonging to the two clusters. Therefore, $C_0$ will be linked to both $C_1$ and $C_2$ (the two points $p_0$ and $p_1$ belong to $C_0$ and they are both linked to points belonging to $C_1$ and $C_2$). On the opposite, $C_3$ does not receive any link from the past, because none of its points receive a point--to--point link.}
\label{fig:dynamic_linking}
\end{figure}

Once all the $3D$ clusters of points are created at each instant of time, we need to dynamically link points at subsequent instants of time, actually building the $(3D+1)$ graph. We have to be careful doing this, because missing dynamical links may result in fragmented trajectories, while extra--links increase the connectivity of the graph, creating false occlusions and making the solution of the problem hard. We define point--to--point dynamical links using a dynamical proximity method whose only assumption is that each $3D$ point moves with a constant velocity between two consecutive instants of time, see Fig.~\ref{fig:dynamic_linking}. Note that the constant velocity assumption is reasonable when working on data collected at a high frame--rate, as the ones used to test SpaRTA, but it may need some refinements in a more general cases. Once we have built point--to--point dynamical links, we use them to define cluster--to--cluster dynamical links: two clusters $C_1$ and $C_2$ are connected in time if there exists at least one point--to--point link between a point $p_1\in C_1$ and a point $p_2\in C_2$, see Fig.~\ref{fig:dynamic_linking}. For the sake of clarity, we omit here to describe all the details of the temporal linking procedure and we refer the interested reader to Appendix~D.

\subsection{Tackling the occlusions}\label{sec:occlusions_solution}

In this section we analyze in detail how we solve the ambiguous connected components, i.e. the occlusions. Here is the core of the method, which overcomes the occlusions yet keeping the identities of the objects involved.

\begin{figure}[ht]
\begin{center}
\includegraphics[width=1.0\linewidth]{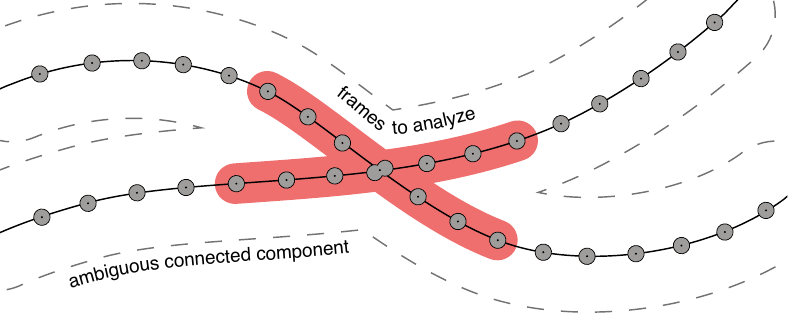}
\end{center}
\caption{\textbf{\textit{3D} occlusion.} A $X$--shape connected component representing a $3D$ occlusion. The two objects are well--separated for the most part of the event, but they share the same cluster during the occlusion. The $4$ branches of the $X$ represent the two different trajectories of the two objects before and after the occlusion, which is represented as a double grey circle at the center of the $X$. During the SDP procedure the analysis of the ambiguous component is restricted to a quite short interval, highlighted in pink.}
\label{fig:x_shape}
\end{figure}

For the sake of simplicity we restrict our attention only on connected components made of two objects in a $3D$ occlusion for one or more frames. The more general case of more than two trajectories belonging to the same connected components can be reduced to the simplest one solving each $3D$ occlusion in a restricted frame range such that only two objects at the time are involved (under the mild assumption that no more than two objects can be involved in the same $3D$ occlusion at the same time).

In an ambiguous connected component, the trajectories of the objects (involved in a $3D$ occlusion) are well--separated for the most part of the event, sharing only few clusters, just during the occlusion. Therefore an ambiguous connected component due to a $3D$ occlusion has the $X$--shape shown in Fig.~\ref{fig:x_shape}, with the $4$ branches of the $X$ representing the trajectories of the two objects before and after the occlusion, which is instead the centre of the $X$. The two occluded objects are not distinguishable and they are detected as one cluster only. The goal of this step is then to identify and separate the volumes occupied by the two objects during the occlusion, \textit{i.e.} to split the clusters in the two subsets representing the volumes of the two distinct objects.

To handle this situation we switch back from clusters to $3D$ cloud of points, representing the ambiguous component as a graph with its $3D$ points as nodes connected by links carrying \textit{static} (equal time) and \textit{dynamic} information (consecutive frames). Following the literature about graph bi--partitioning techniques \cite{boykov2004experimental}, we address the partitioning as an energy minimization problem. Therefore, we associate an energy, $H$, defined as follows:

\begin{equation}\label{eq:H_SDP}
  H = - \sum_{i,j} w_{ij}x_ix_j
\end{equation}

\noindent
where $i$ and $j$ are two different points of the graph, $x_i=\pm 1$  identifies in which partition the point $i$ belongs and $w_{ij}$ is the a coefficient associated to the pair of points $i$ and $j$: because we want to minimize $H$, and it has a minus in front of the sum, $w_{ij}$ will be larger the higher the probability that $i$ and $j$ belong to the same partition. Clearly, a sensible definition of $w_{ij}$ is of paramount importance and some heuristics is inevitable in the choice. Despite this, there is a general principle that has been key to solve the problem: not only we must use a {\it positive} coefficient $w_{ij}$ when it is highly likely that $i$ and $j$ belong to the same partition, but it is also essential to assign a {\it negative} coefficient when it is likely that $i$ and $j$ belong to different partitions. Finally, it is reasonable to have a zero value of $w_{ij}$ whenever it is unclear what is the likely fate of $i$ and $j$. To implement this scenario, we use the simplest rule, which amounts to link the coefficients $w_{ij}$ to the spatio-temporal proximity of the points $i$ and $j$,

\subsubsection{Static coefficients}
We first define the coefficients between points at the same instant of time. We \textit{statically} connect two points, $i$ and $j$, at a mutual distance $d_{ij}$, with the following weight:

\begin{equation}\label{eq::wij}
  w_{ij}(t) = e^{-(d_{ij}/r_1)^\beta} - \, \left(\frac{d_{ij}-r_0}{r_1}\right)^2 \ \theta(d_{ij}-r_0)
\end{equation}

\noindent
$r_1$ is the median nearest neighbor distance, which is also the only natural unit of length of the system of points, hence all other lengths will be measured in units of $r_1$ to make all coefficients dimensionless;  $r_0$, on the other hand, is the
 
\begin{figure}[h!]
\begin{center}
\includegraphics[width=0.9\linewidth]{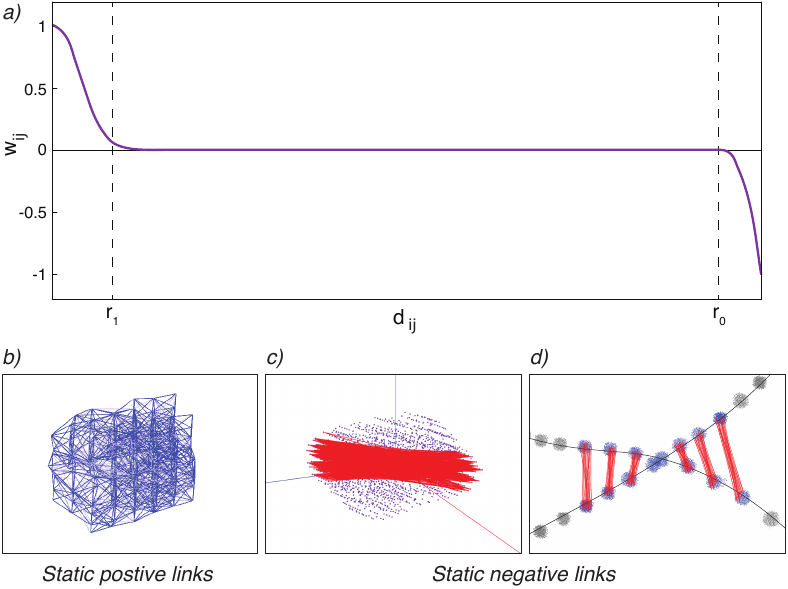}
\end{center}
\caption{\textbf{\textit{3D} occlusion: static linking.} \textbf{a.} The static coefficient, $w_{ij}$ between two points $i$ and $j$ as a function of their mutual distance $d_{ij}$, with $r_0$ and $r_1$ of the specific $3D$ occlusion shown in Fig.~\ref{fig:3D_split}. \textbf{b-d.} Points belonging to the same frame are strongly connected if they are at a very short mutual distance (\textit{static} positive links), while they are strongly disjointed, through a high but negative weight, when at a large mutual distance, both within the same cluster (\textit{static} negative links) or in two different clusters (\textit{static} negative links between clusters).}
\label{fig:sdp_links}
\end{figure}

\noindent 
median size of the reconstructed clusters at that specific instant of time, $r_0\gg r_1$;  $\theta(x)$ is the Heaviside function. The actual shape of  $w_{ij}$ in Eq.~(\ref{eq::wij}) as a function of the space distance $d_{ij}$ is depicted in Fig.\ref{fig:sdp_links}: the idea is to strongly and positively connect those points which are likely to belong to the same cluster, i.e. with a mutual distance smaller than $r_1$, and to negatively connect those points which are likely to belong to different clusters, i.e. at a mutual distance bigger than the usual size of the objects; finally, all points with an uncertain distance, namely $r_1 < d_{ij} < r_0$, have a non-committal $w_{ij} \sim 0$. Hence, the exponent $\beta$ simply rules how sharp is the elbow around $r_1$, and its value does not impact significantly on the results as long as $\beta \geq 2$ (we use $\beta=2.2)$; note, though, that any other sharp decay would do the job.

\subsubsection{Dynamic coefficients}
Using an identical philosophy, we \textit{dynamically} connect two points, $i$ at time $t$ and $j$ at time $t+1$, with the following weight:

\begin{equation}
  w_{ij}(t,t+1) = e^{-D_{ij}/r_1}
  \label{wdinamico}
\end{equation}

where,

\begin{equation}
  D_{ij}(t,t+1) =|\vec r_i(t)+\Delta\vec r_i(t,t+1) - \vec r_j(t+1)|
\end{equation}

\medskip

is the distance between the extrapolated position of $i$ at time $(t+1)$, namely $\vec r_i(t)+\Delta\vec r_i(t,t+1)$, and the position of point $j$ at time $t+1$. The displacement $\Delta\vec r_i(t,t+1)$ is linearly extrapolated from past frames under the assumption that each point moves at a constant velocity between two consecutive frames.
Notice that once again we have used the median nearest neighbour distance $r_1$ as the natural length scale of the system to make the coefficient dimensionless. The meaning of these links is to strongly connect, in time, those pairs of points belonging to  the same dynamic cluster, namely the points belonging to the same branch of the ambiguous $X$ shape (see Figs.~\ref{fig:x_shape} and \ref{fig:3D_split}), which are likely to belong to the same partition.

\subsubsection{Graph partitioning}

In order to find the partitioning which minimizes the energy $H$ in Eq.~\ref{eq:H_SDP} one could use standard approaches, such as integer linear programming or Montecarlo techniques; these, however, are known to fail to find the correct solution (ground state) when there are many local minima of $H$, which is the case of the present problem. Therefore, we choose to approach the problem by using a more robust algorithm based on Semi-Definite Programming (SDP), whose details can be found in \cite{ricci2016performance}, and which successfully finds the absolute minimum of complex energy functions even in presence of multi-minima landscapes.

\begin{figure}[h!]
\begin{center}
\includegraphics[width=1.0\linewidth]{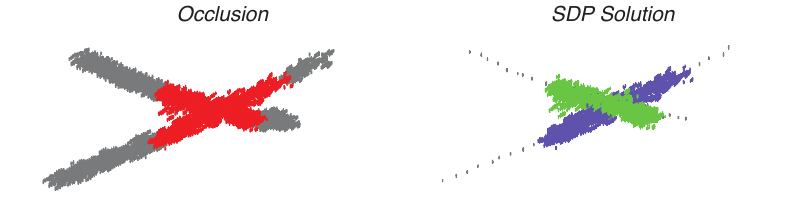}
\end{center}
\caption{\textbf{\textit{3D} occlusion solution.} An example of a $3D$ occlusion from the \textit{Davis-08 dense} dataset \cite{wu2014thermal} used to test the method. On the left: the $X$-shape component, with the $3D$ occlusion and the sub--cloud analyzed with the SDP technique highlighted in red. On the right: the same component after the solution of the $3D$ occlusion, with the two separated trajectories highlighted in blue and green. In the inset: the clusters of the two different targets once the occlusion is solved.}
\label{fig:3D_split}
\end{figure}

Once the $3D$ occlusion is solved through the SDP procedure, the ambiguous $X$--shaped component is divided into two partitions, connected both in space and in time, as shown in Fig.~\ref{fig:3D_split}. The occluded potential clusters are actually split into two sub--clusters representing the volumes occupied by the two objects during the occlusion. In this way, we successfully solve the occlusion retrieving the two trajectories and maintaining the identities during the entire event. Ideally the optimization should be applied to the entire X--shape component, but this would require high computational resources and high computational time, since the minimization of the energy $H$ in Eq.(\ref{eq:H_SDP}) is an NP--hard problem. Hence, in order to reduce the number of variables of the problem, we restrict the optimization to a shorter interval of time, namely from few frames ($3$ in the specific case of the tests presented in Section \ref{sec:results}) before the occlusion starts, to few frames after the occlusion ends, as shown in Fig.~\ref{fig:x_shape} and Fig.~\ref{fig:3D_split} (left panel)  where the interval to be analyzed is highlighted in red.

\section{Experiments and discussion}\label{sec:results}

We tested the method on two public datasets \cite{wu2014thermal} of Brazilian bats colonies emerging from a natural cave in Texas (see Fig.\ref{fig:dataset}), acquired with a system of three synchronized high--speed cameras. To the best of our knowledge these are the only public $3D$ datasets of featureless objects, but they are not in the form of $(3D+1)$ cloud of points. Therefore SpaRTA cannot be directly applied to the data, but a pre--processing procedure on the images is needed. We used standard computer vision techniques to detect the targets of interest in the images and to reconstruct the corresponding $3D$ clouds of points matching the information across the three cameras, see Appendix~C and Appendix~E for a detailed description of the pre--processing procedure and refinements of the method needed when working on multicamera data.

\subsection{Evaluation metrics}

The set of trajectories retrieved by SpaRTA, see Fig.\ref{fig:dataset}, is compared with the set of groudtruth trajectories (published together with the two datasets in \cite{wu2014thermal}) and its quality is assessed using the standard CLEAR MOT evaluation method proposed by K. Bernardin and R. Stiefelhagen in \cite{bernardin2008evaluating} and the metrics described in \cite{wu2014thermal}. The most important metrics in this context is the so-called  Multiple Object Tracking Accuracy parameter (MOTA); this quantity combines three observables, namely: i) the number of false--positive objects (reconstructed objects not belonging to the groundtruth set); ii) the number of missing objects (objects belonging to the groundtruth but not reconstructed by the algorithm); iii) the number of identity--switches (reconstructed trajectories switching between two different groundtruth identities). 

\begin{savenotes}
\begin{table*}[h!]
    \centering
        \begin{tabular}{clcccccc}
            \toprule
            Dataset           & Algorithm &   Class & \textbf{MOTA} &   \textbf{IDS}  & MT      &   ML   &   FM    \\
                              &           &         & ($\%$)  & ($\#$) & ($\%$) & ($\%$) &    ($\#$)       \\
            \midrule
            \midrule \\
                             & \textbf{SpaRTA}    & \textbf{COP-RT} & ~$\mathbf{87.4}$ & ~$\mathbf{25}$  & $\mathbf{92.8}$  & $\mathbf{0.5}$  & $\mathbf{258}$            \\
                             & MHT         & SIP-RT & ~$64.1$ & ~$97$  & $96.6$  & ~~$0$  & $105$            \\
            \textit{Davis--08 sparse}  & SDD-MHT    & SIP-RT & ~$78.9$ & $126$  &  $95.2$  & ~~$0$  & $145$           \\
                             & CP(LDQD)  & SIP-TR & ~$88.1$ & $126$  & $97.1$  & ~~$0$  & $115$            \\
                             & GReTA\textsuperscript{*}
                             & SIP-TR & ~$83.1$ & ~$9$  & $85.1$  & $1.9$  & $167$      \\
                             \\

            \hline                 
            \\
                             & \textbf{SpaRTA}    & \textbf{COP-RT} & ~$\mathbf{86.3}$ & ~$\mathbf{19}$  & $\mathbf{83.7}$  & $\mathbf{6.9}$  & $\mathbf{265}$           \\
                             & MHT   & SIP-RT  & $-32.0$ & $355$     & $71.9$  & $2.5$  & $274$            \\
            \textit{Davis--08 dense~}  & SDD-MHT  & SIP-RT & ~$44.9$ & $444$   & $61.1$  & $3.0$  & $454$                  \\
                             & CP(LDQD) & SIP-TR & ~$80.5$ & $156$  & $84.2$  & $0.5$  & $176$        \\
                             & GReTA\textsuperscript{*} & SIP-TR & ~$79.4$  & ~~$7$     & $80.3$  & $3.9$  & $358$                  \\\\
            \bottomrule
            \\
        \end{tabular}
        \caption{Comparison of the quality of the trajectories retrieved by SpaRTA and the algorithms: MHT, SDD-MHT \cite{wu2014thermal}, CP(LDQD) \cite{wu2016global} and GReTA \cite{attanasi2015GReTA} on the public datasets labeled \textit{Davis-08 sparse} and \textit{Davis-08 dense} published in \cite{wu2014thermal}. In the Table we report the MOTA (Multiple Object Tracking Accuracy) and also: the number of switches of identities (IDS), the percentage of mostly tracked (MT) and mostly lost (ML) trajectories corresponding to groundtruth trajectories which are correctly reconstructed respectively for more than the $80\%$ and for less than the $20\%$ of their time length, the number of tracks fragments (FM) corresponding to the number of times that a groundtruth trajectory, correctly reconstructed, is interrupted. A perfect tracking algorithm produces MOTA~$=100\%$, IDS~$=0$, MT~$=100\%$, ML~$=0\%$, FM~$=0$. In order to compute a match between groundtruth and reconstructed trajectories we chose a miss/hit threshold equal to 0.3m as suggested in \cite{wu2014thermal}. \newline GReTA\textsuperscript{*}: The results obtained by GReTA presented in this table are not the ones published in \cite{attanasi2015GReTA}. This is because, performing the quality evaluation of our new algorithm SpaRTA, we found a shift of one frame in the annotated file of the dataset published in \cite{wu2014thermal}. We evaluated SpaRTA using the annotated file, but taking care of the time shift and for coherency we also updated the results obtained by GReTA.}\label{tab:wu}
\end{table*}
\end{savenotes}

MOTA can be interpreted as the fraction of groundtruth objects correctly reconstructed by the tracking method; its ideal value is equal to $100\%$ (note, however, that -- weirdly as it may seem -- MOTA can have negative values when the number of false--positive plus miss--reconstructed objects plus identity switches exceeds the number of groundtruth objects, quite clearly a rather bad scenario). The second most important metrics is IDS (Identity Switches), which identifies how many times the identities of different actual targets are switched, this is also a rather crucial parameter, as (the consequences of IDS in data analysis are often the most severe).

The other evaluation parameters are the following: MT (Mostly tracked) -- fraction of groundtruth trajectories correctly reconstructed for more than the $80\%$ of their time length; ML (Mostly Lost) -- fraction of groundtruth trajectories that are correctly reconstructed, but for less than the $20\%$ of their time length; FM (Fragmentation) -- corresponding to the number of times that a groundtruth trajectory, correctly reconstructed, is interrupted. It was not possible to evaluate the last parameter MOTP (Multiple Object Tracking Precision), which measures the average distance in the $3D$ space between the groundtruth and the reconstructed objects, because the dataset do not give the actual $3D$ positions of the targets but their estimates based on the SIP approach used by the author in \cite{wu2014thermal}. Hence,  with the MOTP we would have not evaluated the precision of our method, but only the difference between the $3D$ positions obtained by SpaRTA with the ones obtained in  \cite{wu2014thermal}.

\subsection{Results}

\begin{figure}[h!]
\begin{center}
\includegraphics[width=1.0\linewidth]{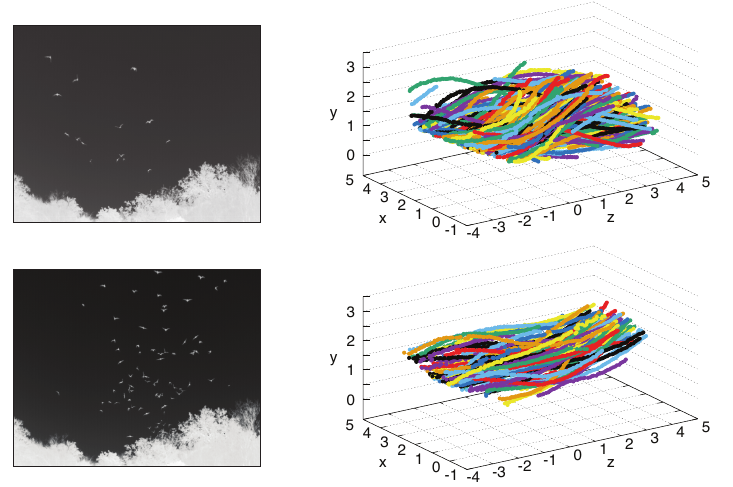}
\end{center}
\caption{\textbf{Dataset and results.} First row: on the left an image from the \textit{Davis-08 sparse} dataset and on the right a sample of the $3D$ trajectories, each with a different color, of the same dataset reconstructed by SpaRTA. Second row: on the left an image from the \textit{Davis-08 dense} dataset and on the right a sample of the $3D$ trajectories reconstructed by SpaRTA}
\label{fig:dataset}
\end{figure}

The performance of SpaRTA is reported in Table~\ref{tab:wu} and compared with four other methods: MHT \cite{wu2014thermal}, SDD--MHT \cite{wu2014thermal}, CP(LDQD) \cite{wu2016global} and GReTA \cite{attanasi2015GReTA}. The difference between the two datasets is that one of them is {\it sparse}, and the data sequence is rather long ($1100$ frames), while the second dataset is {\it dense}, with a far shorter sequence ($200$ frames), see Fig.\ref{fig:dataset}. The tests made on SpaRTA were run on a laptop equipped with a 3.3 GHz i7 Intel processor and with 16 GB of RAM. Moreover, SpaRTA is implemented in C++ using the OpenCV library \cite{opencv_library}.

The comparison with the state--of--the--art methods shows the high performance of SpaRTA, especially in terms of tracking accuracy (MOTA) and number of identity switches (IDS), with satisfactory values of mostly tracked (MT) and mostly lost (ML) groudtruth trajectories, but with a high value of fragmentation (FM) \footnote{The high number of FM in SpaRTA is mainly due to miss--detection of the targets on the images during the pre--processing procedure, which was not optimized for this dataset. When a miss-detection occurs the corresponding $3D$ object cannot be reconstructed, causing the interruption of the correspondent trajectory. This issue can be improved by developing a detection algorithm designed for this dataset; this, however, goes beyond the scope of the present work.}.

On the sparse dataset, SpaRTA outperforms the two SIP--RT methods (MHT and SDD-MHT) in terms of MOTA and IDS, but with a lower number of MT and higher values of ML and FM. Hence, the set of trajectories retrieved by SpaRTA offers a lower coverage of the groundtruth set, but it is better from a qualitatively point of view providing a more accurate set of trajectories for the data analysis, thanks to the high value of MOTA and low number of IDS and consequently a negligible number of wrong trajectories. Moreover, the comparison of SpaRTA with the two SIP-TR methods, CP(LDQD) and GReTA, shows that the performance of the three methods are comparable, with SpaRTA offering the best balance between the value of MOTA and the number of IDS: CP(LDQD) produces the highest value of MOTA ($88.1\%$), comparable with the one obtained by SpaRTA ($87.4\%$), but with a higher number of IDS, while GReTA produces the lowest number of IDS ($9$), comparable with the ones produced by SpaRTA ($25$), but with a lower value of the MOTA parameter.

On the dense dataset, SpaRTA exhibits the best performance in terms of MOTA ($86.3\%$), even though with the highest number of ML \footnote{The high number ML is a consequence of the formation of  "ghost" trajectories (trajectories not representing existing targets), intrinsic to all RT methods as explained in detail in Appendix E. In SpaRTA these trajectories are identified from their time length and removed from the retrieved set of trajectories, so that short groundtruth trajectories may be mislead for ghost and for this reason thrown away.}. Despite this high value of ML, SpaRTA shows an excellent result in terms of MOTA, particularly outstanding when compared with the two SIP--RT methods that produce low value of MOTA, even negative in the case of the MHT algorithm, thus  implying that the number of false positive and missing reconstructing objects exceed the number of correctly reconstructed targets. Moreover the low number of IDS ($19$), second only to GReTA, together with the high fraction of mostly tracked trajectories ($83.7\%$) comparable with the highest ratio of MT achieved by CP(LDQD), confirms the high--performance of our method that offers the best balance between the evaluation parameters, as it happens for the sparse dataset.

\section{Conclusions}

We proposed a tracking method, SpaRTA (Spatiotemporal Reconstruction Tracking Algorithm), designed to track large and dense group of featureless objects, without any specific prerequisites on the $3D$ system used to acquired the data. SpaRTA works with $(3D+1)$ dense clouds of points representing the volumes occupied in time by the targets of interest. Each cloud of points is partitioned in spatio-temporal connected components, corresponding to the trajectories of single individuals in the group. The method is designed to handle efficiently the ambiguities stemming from occlusions, i.e. objects getting too close in the $3D$ space to be detected as separated entities; to this aim SpaRTA employs an optimization routine inspired on Semi--Definite Programming (SDP) techniques introduced in the field of statistical mechanics \cite{ricci2016performance}. Apart from using in a proficous way for the first time SDP in the computer vision context, the true core of SpaRTA is the novel way in which the spatio-temporal graph is built: the key idea is to define an energy (or cost) function based on the use of both {\it attractive} and {\it repulsive} links between points within the cloud, in such a way to separate with relatively little numerical effort the ambiguous cases by minimizing such energy.

SpaRTA was tested on two public datasets, \cite{wu2014thermal}, producing high quality results in terms of correctness of the trajectories, evaluated through the standard quality parameter MOTA, and producing a low rate of identity switches. The retrieved trajectories were compared with the four state--of--the--art tracking methods MHT \cite{wu2014thermal}, SDD-MHT \cite{wu2014thermal}, CP(LDQD) \cite{wu2016global} and GReTA \cite{attanasi2015GReTA}. The greatest advantage of SpaRTA over the other methods is an outstanding MOTA, combined with an excellent IDS (second only to GReTA) and a very good MT. This means that, not only SpaRTA is able to achieve a quantitatively satisfying coverage of the set of trajectories, but it does that with the lowest number of false positives and a remarkably low number of identity switches compared to most other methods. At the level of data analysis, this kind of performance guarantees that the completeness of the data coverage is not jeopardized by a qualitative disruption of the results, due to the severe consequences of having wrong trajectories
.

\section*{Acknowledgments}

This work was partly supported by  European Research Council, Proof of Concept Grant n. 713651. We thank I. Giardina and M. Viale for the advice and the fruitful discussion on the new tracking strategy.

\newpage

\appendices


\section{Multicamera data: SIP vs COP approach}

When working with multicamera data, there are two fundamentally different ways to represent the targets an algorithm wants to track: on one hand Single Point (SIP) representation, where each target at a certain instant of time is associated to one single identity and spatial position (typically the baricenter); on the other hand dense Cloud of Points (COP) representation, where each target at a certain instant of time is associated to a spatial volume.

\begin{figure}[h!]
\begin{center}
\includegraphics[width=1.0\linewidth]{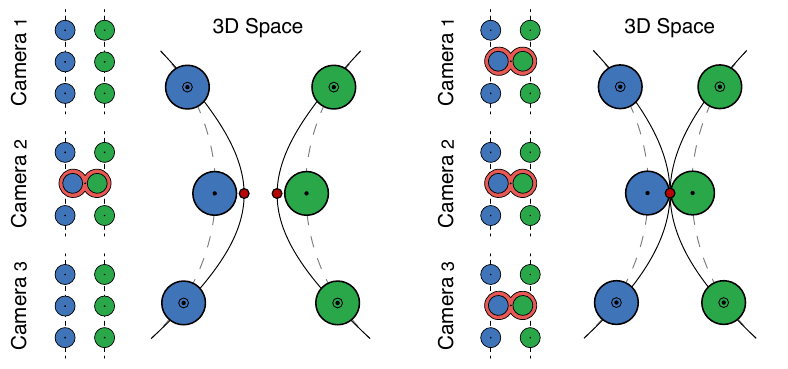}
\end{center}
\caption{\textbf{Standard \textbf{3D} reconstruction of occluded targets.} On the left a $2D$ occlusion: the blue and the green targets are occluded and detected as only one object (highlighted with a red contour in the scheme), in Camera~2 where they are both associated with the same $2D$ point (black dot at the centre of the detected object), while they are associated to different $2D$ points in the other two cameras. At the time of the occlusion, the two targets are reconstructed in the $3D$ space as two different $3D$ points (red circles), but the two $3D$ positions do not correspond to the \textit{real} baricenters of the two targets. As a consequence, the retrieved trajectories (dashed black lines) do not produce an accurate approximation of the \textit{real} ones (solid black lines). On the right a $3D$ occlusion: the blue and the green targets are occluded in all the cameras (highlighted with a red contour in the schemes of the three cameras). In each camera, the two targets are detected as only one object (black dots in the scheme) and they are reconstructed with the same $3D$ position (red circle), which lies between the two \textit{real} baricenters. The two targets are then not separated in the $3D$ space and the two reconstructed trajectories (dashed lines) are not an accurate approximation of the two \textit{real} trajectories (solid lines).}
\label{fig:standardVsSparta}
\end{figure}

In the SIP state--of--the--art multicamera tracking methods, targets of interest are detected in the images through a segmentation routine and they associate to each detected target a single $2D$ coordinate, which is generally the target baricenter on the images. Regardless the approach used for tracking, target $2D$ baricenters need to be matched across cameras to retrieve the correspondent $3D$ point, which in principle should be the $3D$ baricenter of the target.

The assumption behind this procedure is that the $3D$ baricenter of an object corresponds to the $2D$ baricenters of its images, which is reasonable when dealing with not--occluded targets but not in a general case where $2D$ and $3D$ occlusions occur. Indeed, as shown in the left panel of Fig.~\ref{fig:standardVsSparta}, when $2D$ targets (the green and the blue) are occluded in one camera only, the occluded objects are associated to the same $2D$ position in the camera where the occlusion occurs, while they correspond to different coordinates in the other cameras. The information from the two cameras where the occlusion do not occur make the occluded targets to be reconstructed in two different positions (highlighted with red dots in the figure), but with a poor accuracy: both the two $3D$ positions do not correspond to the target baricenters. The COP representation, on the other hand, allows to associate to each object (at each instant of time) a dense cloud of $3D$ points and not only its baricenter. This dense representation actually reconstruct the volumes occupied by the detected objects, even during a $2D$ occlusion; their $3D$ positions are computed using the entire cloud and they actually correspond to the $3D$ baricenters of the two separated objects, thus giving a better approximation of the \textit{real} position of the target.

A more severe limitation of SIP representation concerns the solution of $3D$ occlusions. In this particular situation two objects are occluded in all the cameras, as shown in the right panel of Fig.~\ref{fig:standardVsSparta}. Indeed the association of each detected target with only one $3D$ position (the red dot in Fig.~\ref{fig:standardVsSparta}) implies that the same $3D$ coordinate is associated to multiple targets during the occlusion, with the consequent loss of identity for all the targets involved. Instead in the COP representation, when two or more targets get in a $3D$ occlusion, their total volume may be split in sub--parts each corresponding to the volume of one single object, as SpaRTA does with the use of the SDP procedure described in Section~4.2 of the manuscript. Therefore targets identities are not lost anymore and the $3D$ baricenters of each target may then be computed on the correspondent sub--cloud with high--accuracy.

Trajectories retrieved with SIP representation are then less accurate in the computation of the positions of the targets (in the case of both $2D$ and $3D$ occlusions) than trajectories retrieved with COP representation and they are also more prone to identity switches after a $3D$ occlusion occurs because occluded targets are reconstructed as only one single $3D$ object.

\section{Multicamera data: Occlusions}

One of the most spread experimental method for $3D$ tracking is represented by acquire data with a system of multicamera, namely two or more cameras shooting synchronously images of the same group of moving targets. 

\begin{figure*}[!h]
\begin{center}
\includegraphics[width=1.0\linewidth]{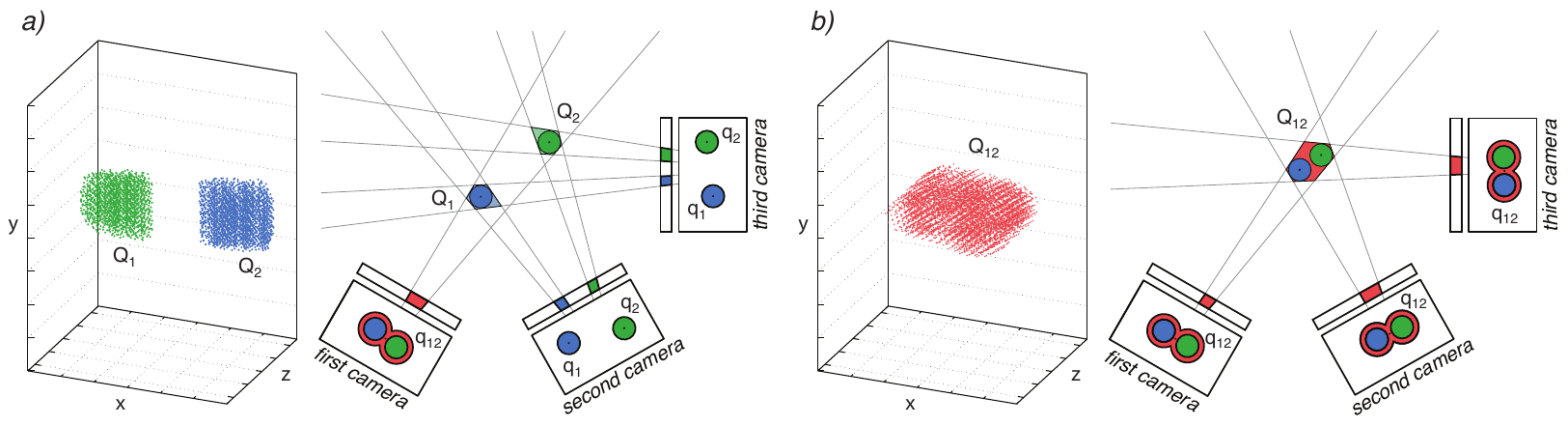}
\end{center}
\caption{\textbf{Optical occlusions.} a) $2D$ occlusions: two objects are well--separated in the $3D$ space (the blue and the green clouds of points) but occluded in one of the three cameras and they are reconstructed as two well--separated clouds of points. b) $3D$ occlusions: two objects are in $3D$ proximity and therefore occluded in all the three cameras, so that they are reconstructed as only one cloud of points.}
\label{fig:occlusion}
\end{figure*}

The main issue of all tracking methods is how to handle occlusions, which arise everytime that two or more objects get too close to be detected as multiple identities. In the specific case of data collected with multicamera systems two different kind of optical occlusions may occur: \textbf{2d--occlusion} which arise when two or more objects, separated in the $3D$ space, are too close to be individually detected in the $2D$ space of one camera only, and \textbf{3d--occlusions} which arise when multiple objects are occluded in all the cameras simultaneously, namely when multiple objects get in $3D$ proximity, as shown in Fig.~\ref{fig:occlusion}. 

As outlined in Section~2 of the manuscript, state--of--the--art tracking approaches on data collected via multicamera systems may be divided into two main classes: tracking--reconstruction (TR) and reconstruction--tracking (RT) methods. TR algorithms address the problem in the $2D$ space of the cameras: objects are first tracked in each camera, and $2D$ trajectories are then matched across the cameras to retrieve the corresponding $3D$ trajectories, hence they need to handle both $2D$ and $3D$ occlusions. $2D$ occlusions make the tracking in the $2D$ space of the cameras ambiguous and all the TR algorithms have then to be focused on how to cope with this kind of occlusions, solving the $2D$ ambiguities and at the same time keeping the computational resources at an acceptable level. On the other hand, RT methods, to which SpaRTA belongs when applied to multicamera data, address $3D$ tracking directly in the $3D$ space. Therefore $2D$ occlusions are naturally avoided, because they use the information of the cameras where the $2D$ occluded targets are well--separated, and they have to deal only with $3D$ occlusions.

A better explanation of the conceptual difference between $2D$ and $3D$ occlusions is shown in Fig.~\ref{fig:occlusion}: in panel a) two targets (the green and the blue) well--separated in $3D$ space are detected as only one object in the $2D$ plane of the first camera, while they are well--separated in the other two cameras. The use of the information of the three cameras guarantees that they are reconstructed as two  different and separated $3D$ objects. In panel b) the two targets are, instead, seen as only one object in all the three cameras, their identities are not distinguishable anymore (because none of the cameras has the information on their multiple identities) and they are reconstructed as only one $3D$ object, highlighted in red.

When applied to data collected with a multicamera system, as in the tests presented in Section~5 of the manuscript, SpaRTA follows an RT approach since it addresses the tracking problem directly in the $3D$ space and thus, as all the RT methods, it is only affected by $3D$ occlusions.

\begin{figure*}[h!]
\begin{center}
\includegraphics[width=1.0\linewidth]{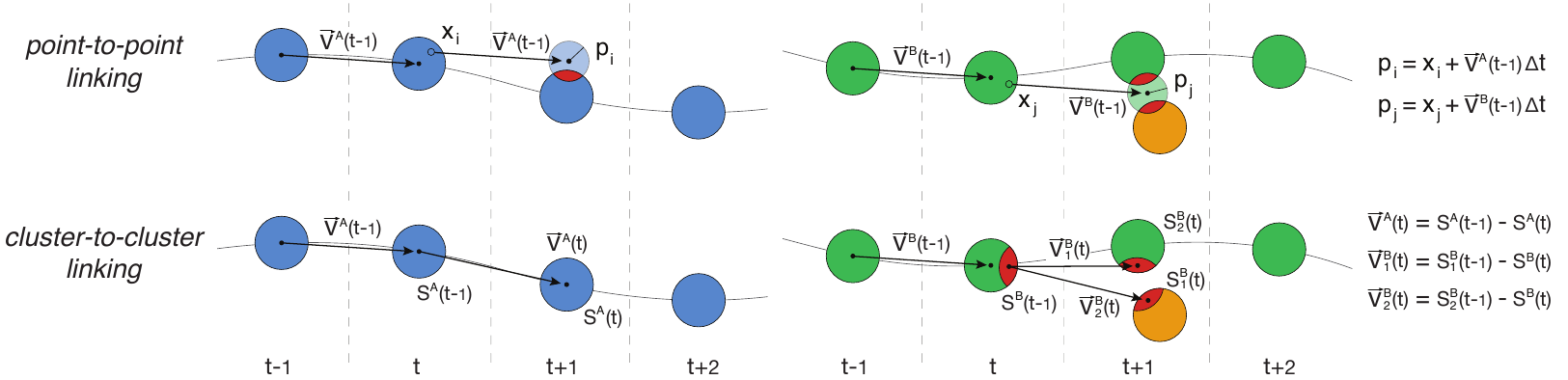}
\end{center}
\caption{\textbf{Cluster graph construction: dynamic linking.} Blue, green and orange circles represent reconstructed clusters at different instant of time. At a generic frame $t$, a point--to--point linking procedure is performed. Point--to--point links are then used to define cluster--to--cluster links. \textbf{Point--to--point linking:} given the position of a point $x$ at frame $t$, its position at the next frame, $p$, is computed using the velocity vector associated to the link received from the past by the cluster to which $x$ belongs. $x$ is then connected to all the $3D$ points at a distance from $p$ smaller than a certain threshold. In the example on the left, $x_i$ is connected to all the points in the region highlighted in red, which belong to the same cluster. In the example on the right, $x_j$ is connected with all the points in the two red regions, belonging to two different clusters. C\textbf{luster--to--cluster linking:} two clusters are connected if there exists at least one point--to--point link between points belonging to the two clusters. In the example on the left, the blue cluster at frame $t$ is connected only with the blue cluster at frame $t+1$. The velocity vector associated to this link is then computed as the spatial displacement between the baricenters of the two clusters. In the example on the right, instead, the green cluster at time $t$ is connected with both the green and the orange cluster at frame $t+1$. The velocity vectors associated to these links are computed as the spatial displacement between the baricenter of the sub--clusters highlighted in red. This refinement guarantees high quality of the dynamic links even when an occlusion occurs.}
\label{fig:dynamicLinking}
\end{figure*}

\section{From stereo--images to \textit{(3D+1)} cloud of points}\label{sec:stereo2clouds}

In this Appendix we will give the details on the procedure we use to reconstruct the $(3D+1)$ cloud of points from a set of images collected with a system of synchronized high--speed cameras. We applied this procedure to test our new algorithm, SpaRTA, on the public benchmarks in \cite{wu2014thermal}, which are collected with three synchronized cameras. For this reason in this Appendix we will refer to systems of three synchronized cameras, but the entire procedure may be easily generalized to any multicamera system.

\medskip

In order to obtain the $(3D+1)$ cloud of points from the images, we perform the following steps:

\medskip

\textbf{1~-- Segmentation.} The goal of the segmentation routine is to identify the \textit{active} pixels in the images, \textit{i.e.} pixels representing the objects to be tracked. Therefore this step strictly depends on the objects appearance and on the camera system used. The results of the entire tracking algorithm may be heavily affected by segmentation errors, since miss--detected as well as false detected objects may lead to fragmented or, even worst, to \textit{ghost} trajectories, \textit{i.e.} not corresponding to any real object. Therefore, it is necessary to design the segmentation routine for each specific dataset. In this particular case, we designed the segmentation routine as a standard background subtraction over a sliding window as suggested in \cite{sobral2014comprehensive}, since SpaRTA was developed to track featureless objects using monochromatic images, characterized by high color contrast between the moving objects and a still background, \textit{i.e.} dark objects over a light background or vice versa. 
\medskip

\textbf{2~-- Pixels matching.} At each instant of time, the sets of \textit{active} pixels, $P_1$, $P_2$ and $P_3$ (one for each camera) identified at the previous step, are analyzed to find the triplets $\tau=(p_1, p_2, p_3)$ of $2D$ points (pixels) $p_i\in P_i$, projections of the same $3D$ point in the three cameras. A triplet $\tau$ is considered a good match if its reprojection error, see \cite{hartley2003multiple}, is smaller than a threshold, that we choose equal to 1.5px. In principle we should reconstruct and check all the possible $N^3$ triplets, with $N$ being the average number of \textit{active} pixels in each camera. However, using the trifocal constraint \cite{hartley2003multiple} the number of triplets to be checked may be reduced to $O(N)$.

\medskip

\textbf{3~-- \textit{(3D+1)} cloud of points creation.} All the triplets $\tau$ created at the previous step are reconstructed in the $3D$ space, using a standard DLT reconstruction procedure \cite{hartley2003multiple}, thus forming a $(3D+1)$ cloud of points which represents the $3D$ volume occupied by all the targets during the entire event of interest.

\section{Dynamic linking procedure}\label{sec:dynamicLinking}

This procedure is meant to identify and link in time those $3D$ clusters corresponding to the same object at subsequent instants of time, actually building the clusters graph. This is the most delicate step of the method, because it can affect the quality of the result of the entire procedure: missing links may indeed result in fragmented trajectories, while extra--links increase the connectivity of the graph, creating fake occlusions and making the solution of the problem hard. 

\medskip

The idea is to use both spatial and temporal information using the duality between an object as a dense cloud of points and an object as a cluster. Indeed, we first define point--to--point links, using the only assumption that each $3D$ point moves with a constant velocity between two consecutive instants of time.  Point--to--point links are then used to define cluster--to--cluster links: two clusters $C_1$ and $C_2$ are connected in time if there exists at least a point--to--point link between a point $p_1\in C_1$ and a point $p_2\in C_2$, see Fig.~\ref{fig:dynamicLinking}. 

\medskip

The first instant of time, $t=1$, is quite special because we cannot use any dynamic information from the past. Therefore for this special case, we use a procedure based only on the distance between clusters. 

\medskip

In particular, denoting by $C_{i,1}$ with $i=1\cdots,N_1$ the $N_1$ clusters identified at time $t=1$ and by $C_{j,2}$ with $j=1\cdots,N_2$ the $N_2$ clusters identified at time $t=2$:

\begin{enumerate}

    \item we compute the baricenters, $\vec{B}_{i,1}$ and $\vec{B}_{j,2}$, of each cluster $C_{i,1}$ at time $t=1$ and $C_{j,2}$ at time $t=2$, as the average position of the $3D$ points belonging to those clusters;
    
   \item we perform a one--to--one matching, via a standard Munkres algorithm \cite{munkres1957algorithms}, between clusters at time $t=1$ and $t=2$ based on the $3D$ distance between the cluster baricenters. For each matched pair of cluster $(C_{i,1}, C_{j,2})$ we add a link labeled with the vector velocity $\vec{v}$ defined as: 

\begin{equation}
  \vec{v}_{i,1} = \frac{1}{\Delta t}\left(\vec{B}_{j,2}-\vec{B}_{i,1}\right)
\end{equation}

where with $\vec{v}_{i,1}$ we denote the velocity of the object $i$ at frame $t=1$.

\end{enumerate}

\medskip

Instead, for a generic instant of time, $t$, we use spatial information considering the clusters as dense clouds of points and performing at first a point--to--point linking procedure, based on the assumption of constant velocity between two consecutive instants of time. These point--to--point links are then used to define cluster--to--cluster links and therefore to build the graph in Fig.~1 of the main text. 

\medskip

In particular, denoting by $C_{i,t}$ and $C_{j,t+1}$ the clusters identified at time $t$ and $t+1$ respectively, by $\vec{v}_{i,t}$ the velocity of cluster $C_{i,t}$, see Fig.~\ref{fig:dynamicLinking}:

\begin{enumerate}
    \item we compute the predicted position, $\vec{p}_{i,t}$, of each point $\vec{x}_{i,t}\in C_{i,t}$ as:
    \begin{equation}
        \vec{p}_{i,t}=\vec{x}_{i,t}+\vec{v}_{i,t}\Delta t
    \end{equation}
   \item we connect $\vec{x}_{i,t}\in C_{i,t}$ to all those points $x_{j,t+1}\in C_{j,t+1}$, such that:
    \begin{equation}
        d(\vec{x}_{j,t+1}, \vec{p}_{i,t})\leq \bar r
    \end{equation}
    with $\bar r$ chosen as the median nearest neighbor distance, $r_1$.
    \item we define cluster-to cluster links in such a way that two clusters $C_{i,t}$ and $C_{j,t+1}$ are temporally linked if at least one point $x_{i,t}\in C_{i,t}$ is linked with a point $x_{j,t+1}\in C_{j,t+1}$.
\end{enumerate}

\medskip

Once all the clusters with links from the past are analyzed, we consider the set of those clusters without any link from the past at frame $t$ and $t+1$ and we perform a one--to--one match between the two subsets, following the same procedure described above for the first instant of time, $t=1$.

\section{\textit{Ghost} points formation and trajectories removal}\label{sec:ghostPoints}
\begin{figure}[h!]
\begin{center}
\includegraphics[width=1.0\linewidth]{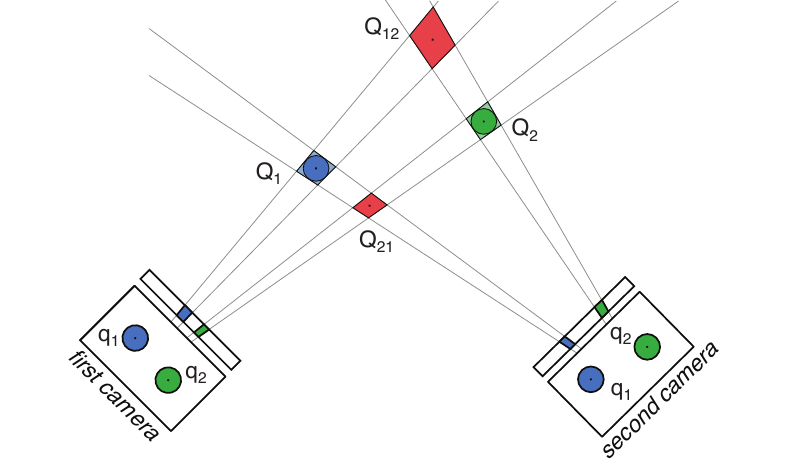}
\end{center}
\caption{\textbf{\textit{Ghost} objects formation.} The blue and the green objects, namely $Q_1$ and $Q_2$ are reprojected in the two cameras as $q_1$ and $q_2$. The pair $(q_1, q_1)$ is a good match because the two pixels are the image of the same $3D$ point $Q_1$, and the pair $(q_2, q_2)$ is a good match as well. The two pairs $(q_1, q_2)$ and $(q_2, q_1)$ do not represent any object of interest but they are good matches from a epipolar perspective, see \cite{hartley2003multiple}. The pixel matching procedure will then reconstruct the two correct objects $Q_1$ and $Q_2$ and the two \textit{ghost} objects $Q_{12}$ and $Q_{21}$. 
The introduction of a third camera drastically reduces the creation of \textit{ghost} points working on triplets of pixels, but even in this case ambiguities are not completely solved.}
\label{fig:ghost_formation}
\end{figure}

The pixel matching procedure described in Appendix~\ref{sec:stereo2clouds} may lead to the formation of \textit{ghost} $3D$ points, representing points not belonging to any \textit{real} $3D$ object. \textit{Ghost} points are an intrinsic artifact of the method which, at this level, cannot discriminate between \textit{ghost} and correct points, as shown in Fig.~\ref{fig:ghost_formation}. Therefore \textit{ghost} $3D$ points are included in the overall $(3D+1)$ cloud of points, creating \textit{ghost} clusters and hence \textit{ghost} trajectories, which has to be identified and removed to not affect the quality of the retrieved solution. 
\begin{figure}[h!]
\begin{center}
\includegraphics[width=1.0\linewidth]{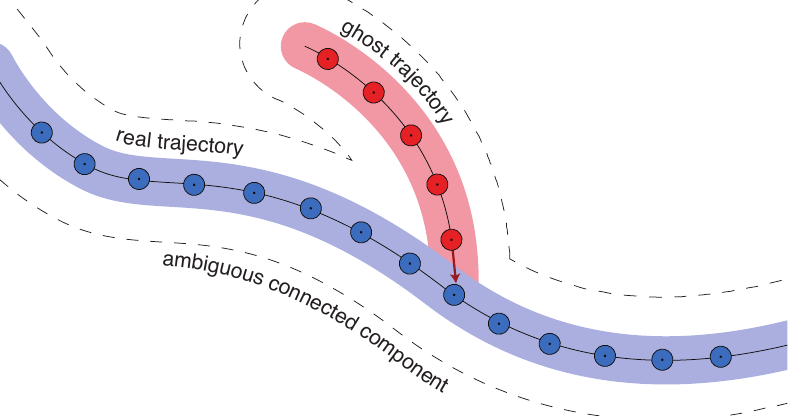}
\end{center}
\caption{\textbf{\textit{Ghost} trajectories removal.} A $Y$--shape connected component representing a real (blue) and a \textit{ghost} (red) trajectories occluded at one frame. This ambiguous component is due to a wrong dynamic link (the red arrow), which can be easily identified from the peculiar $Y$--shape of the component together with the short length of the \textit{ghost} branch. The wrong dynamic link is detected and removed as well as the \textit{ghost} points.}
\label{fig:y_shape}
\end{figure}
\medskip

For their nature, \textit{ghost} trajectories last for very few frames, so that not ambiguous \textit{ghost} connected components can be easily identified and eliminated removing all the trajectories shorter than a certain time length, which has to be empirically chosen depending on the dataset. In the tests presented in Section~5 of the manuscript it is set to $10$.

A more complicated situation occurs when a \textit{ghost} and a real object get in $3D$ proximity creating a ambiguous connected component, with the typical $Y$--shape shown in Fig.~\ref{fig:y_shape} with one of the two branches (the one corresponding to the \textit{ghost} trajectories) quite short. This kind of ambiguity is essentially due to an error in the dynamical linking procedure, which wrongly connects the \textit{ghost} to a real cluster.

We identify the connected components with a $Y$--shape graph and we solve the ambiguity removing the wrong link (highlighted in red in Fig.~\ref{fig:y_shape}) between the \textit{ghost} and the real trajectory, actually breaking the graph into two partitions, representing a long and correct trajectory and a short \textit{ghost} trajectory which is removed.

\bibliographystyle{IEEEtran}
\bibliography{IEEEabrv,biblio}

\begin{thebibliography}{10}
\providecommand{\url}[1]{#1}
\csname url@samestyle\endcsname
\providecommand{\newblock}{\relax}
\providecommand{\bibinfo}[2]{#2}
\providecommand{\BIBentrySTDinterwordspacing}{\spaceskip=0pt\relax}
\providecommand{\BIBentryALTinterwordstretchfactor}{4}
\providecommand{\BIBentryALTinterwordspacing}{\spaceskip=\fontdimen2\font plus
\BIBentryALTinterwordstretchfactor\fontdimen3\font minus
  \fontdimen4\font\relax}
\providecommand{\BIBforeignlanguage}[2]{{%
\expandafter\ifx\csname l@#1\endcsname\relax
\typeout{** WARNING: IEEEtran.bst: No hyphenation pattern has been}%
\typeout{** loaded for the language `#1'. Using the pattern for}%
\typeout{** the default language instead.}%
\else
\language=\csname l@#1\endcsname
\fi
#2}}
\providecommand{\BIBdecl}{\relax}
\BIBdecl

\bibitem{ouellette2006quantitative}
N.~T. Ouellette, H.~Xu, and E.~Bodenschatz, ``A quantitative study of
  three-dimensional lagrangian particle tracking algorithms,''
  \emph{Experiments in Fluids}, vol.~40, no.~2, pp. 301--313, 2006.

\bibitem{wu2016global}
Z.~Wu and M.~Betke, ``Global optimization for coupled detection and data
  association in multiple object tracking,'' \emph{Computer Vision and Image
  Understanding}, vol. 143, pp. 25--37, 2016.

\bibitem{cheng2015novel}
X.~E. Cheng, Z.-M. Qian, S.~H. Wang, N.~Jiang, A.~Guo, and Y.~Q. Chen, ``A
  novel method for tracking individuals of fruit fly swarms flying in a
  laboratory flight arena,'' \emph{PloS one}, vol.~10, 2015.

\bibitem{moussaid2012traffic}
M.~Moussaid, E.~G. Guillot, M.~Moreau, J.~Fehrenbach, O.~Chabiron,
  S.~Lemercier, J.~Pettr{\'e}, C.~Appert-Rolland, P.~Degond, and G.~Theraulaz,
  ``Traffic instabilities in self-organized pedestrian crowds,'' \emph{PLoS
  Comput. Biol}, vol.~8, no.~3, 2012.

\bibitem{wen2016multi}
L.~Wen, Z.~Lei, M.-C. Chang, H.~Qi, and S.~Lyu, ``Multi-camera multi-target
  tracking with space-time-view hyper-graph,'' \emph{International Journal of
  Computer Vision}, pp. 1--21, 2016.

\bibitem{michel2007gpu}
P.~Michel, J.~Chestnutt, S.~Kagami, K.~Nishiwaki, J.~Kuffner, and T.~Kanade,
  ``Gpu-accelerated real-time 3d tracking for humanoid locomotion and stair
  climbing,'' in \emph{Intelligent Robots and Systems, 2007. IEEE/RSJ
  International Conference on}, 2007, pp. 463--469.

\bibitem{ess2010object}
A.~Ess, K.~Schindler, B.~Leibe, and L.~Van~Gool, ``Object detection and
  tracking for autonomous navigation in dynamic environments,'' \emph{The
  International Journal of Robotics Research}, vol.~29, no.~14, pp. 1707--1725,
  2010.

\bibitem{ricci2016performance}
F.~Ricci-Tersenghi, A.~Javanmard, and A.~Montanari, ``Performance of a
  community detection algorithm based on semidefinite programming,'' in
  \emph{Journal of Physics: Conference Series}, vol. 699.\hskip 1em plus 0.5em
  minus 0.4em\relax IOP Publishing, 2016, p. 012015.

\bibitem{wu2014thermal}
Z.~Wu, N.~Fuller, D.~Theriault, and M.~Betke, ``A thermal infrared video
  benchmark for visual analysis,'' in \emph{Proceedings of the IEEE Conference
  on Computer Vision and Pattern Recognition Workshops}, 2014, pp. 201--208.

\bibitem{reid1979algorithm}
D.~Reid, ``An algorithm for tracking multiple targets,'' \emph{IEEE
  transactions on Automatic Control}, vol.~24, no.~6, pp. 843--854, 1979.

\bibitem{attanasi2015GReTA}
A.~Attanasi, A.~Cavagna, L.~Del~Castello, I.~Giardina, A.~Jelic, S.~Melillo,
  L.~Parisi, F.~Pellacini, E.~Shen, E.~Silvestri \emph{et~al.}, ``Greta - a
  novel global and recursive tracking algorithm in three dimensions,''
  \emph{Pattern Analysis and Machine Intelligence, IEEE Transactions on},
  vol.~99, 2015.

\bibitem{cox1996efficient}
I.~J. Cox and S.~L. Hingorani, ``An efficient implementation of reid's multiple
  hypothesis tracking algorithm and its evaluation for the purpose of visual
  tracking,'' \emph{IEEE Transactions on pattern analysis and machine
  intelligence}, vol.~18, no.~2, pp. 138--150, 1996.

\bibitem{wu2011efficient}
Z.~Wu, T.~H. Kunz, and M.~Betke, ``Efficient track linking methods for track
  graphs using network-flow and set-cover techniques,'' in \emph{Computer
  Vision and Pattern Recognition (CVPR), 2011 IEEE Conference on}.\hskip 1em
  plus 0.5em minus 0.4em\relax IEEE, 2011, pp. 1185--1192.

\bibitem{wu2011automated}
H.~S. Wu, Q.~Zhao, D.~Zou, and Y.~Q. Chen, ``Automated 3d trajectory measuring
  of large numbers of moving particles,'' \emph{Optics express}, vol.~19,
  no.~8, pp. 7646--7663, 2011.

\bibitem{liu2012automatic}
Y.~Liu, H.~Li, and Y.~Q. Chen, ``Automatic tracking of a large number of moving
  targets in 3d,'' in \emph{European Conference on Computer Vision}.\hskip 1em
  plus 0.5em minus 0.4em\relax Springer, 2012, pp. 730--742.

\bibitem{tyagi2007fusion}
A.~Tyagi, G.~Potamianos, J.~W. Davis, and S.~M. Chu, ``Fusion of multiple
  camera views for kernel-based 3d tracking,'' in \emph{Motion and Video
  Computing, 2007. IEEE Workshop on}, 2007.

\bibitem{li2002relaxation}
Y.~Li, A.~Hilton, and J.~Illingworth, ``A relaxation algorithm for real-time
  multiple view 3d-tracking,'' \emph{Image and vision computing}, vol.~20,
  no.~12, pp. 841--859, 2002.

\bibitem{dockstader2001multiple}
S.~L. Dockstader and A.~M. Tekalp, ``Multiple camera fusion for multi-object
  tracking,'' in \emph{Multi-Object Tracking, 2001. Proceedings. 2001 IEEE
  Workshop on}.\hskip 1em plus 0.5em minus 0.4em\relax IEEE, 2001, pp. 95--102.

\bibitem{wu2009tracking}
Z.~Wu, N.~I. Hristov, T.~H. Kunz, and M.~Betke, ``Tracking-reconstruction or
  reconstruction-tracking? comparison of two multiple hypothesis tracking
  approaches to interpret 3d object motion from several camera views,'' in
  \emph{Motion and Video Computing, 2009. WMVC'09. Workshop on}.\hskip 1em plus
  0.5em minus 0.4em\relax IEEE, 2009, pp. 1--8.

\bibitem{RGBD-Overview2013}
J.~Han, L.~Shao, D.~Xu, and J.~Shotton, ``Enhanced computer vision with
  microsoft kinect sensor: A review,'' \emph{IEEE Transactions on Cybernetics},
  vol.~43, no.~5, pp. 1318--1334, Oct 2013.

\bibitem{Lidar-Asvadi2016}
A.~Asvadi, P.~Girão, P.~Peixoto, and U.~Nunes, ``3d object tracking using rgb
  and lidar data,'' in \emph{2016 IEEE 19th International Conference on
  Intelligent Transportation Systems (ITSC)}, Nov 2016, pp. 1255--1260.

\bibitem{RADAR-Mobus2003}
R.~Mobus, A.~Joos, and U.~Kolbe, ``Multi-target multi-object radar tracking,''
  in \emph{IEEE IV2003 Intelligent Vehicles Symposium. Proceedings (Cat.
  No.03TH8683)}, June 2003, pp. 489--494.

\bibitem{RGBD-2012}
M.~Munaro, F.~Basso, and E.~Menegatti, ``Tracking people within groups with
  rgb-d data,'' in \emph{2012 IEEE/RSJ International Conference on Intelligent
  Robots and Systems}, Oct 2012, pp. 2101--2107.

\bibitem{ZHANG2013126}
L.~Zhang, Q.~Li, M.~Li, Q.~Mao, and A.~Nüchter, ``Multiple vehicle-like target
  tracking based on the velodyne lidar*,'' \emph{IFAC Proceedings Volumes},
  vol.~46, no.~10, pp. 126 -- 131, 2013, 8th IFAC Symposium on Intelligent
  Autonomous Vehicles.

\bibitem{Lidar-Choi2013}
J.~Choi, S.~Ulbrich, B.~Lichte, and M.~Maurer, ``Multi-target tracking using a
  3d-lidar sensor for autonomous vehicles,'' in \emph{16th International IEEE
  Conference on Intelligent Transportation Systems (ITSC 2013)}, Oct 2013, pp.
  881--886.

\bibitem{RADAR-2004}
R.~Mobus and U.~Kolbe, ``Multi-target multi-object tracking, sensor fusion of
  radar and infrared,'' in \emph{IEEE Intelligent Vehicles Symposium, 2004},
  June 2004, pp. 732--737.

\bibitem{cavagna2016towards}
A.~Cavagna, C.~Creato, L.~Del~Castello, S.~Melillo, L.~Parisi, and M.~Viale,
  ``Towards a tracking algorithm based on the clustering of spatio-temporal
  clouds of points,'' in \emph{VISAPP 2016}, vol.~3, 2016, pp. 679--685.

\bibitem{Rokach2005}
L.~Rokach and O.~Maimon, \emph{Clustering Methods}.\hskip 1em plus 0.5em minus
  0.4em\relax Boston, MA: Springer US, 2005, pp. 321--352.

\bibitem{hopcroft1973algorithm}
J.~Hopcroft and R.~Tarjan, ``Algorithm 447: efficient algorithms for graph
  manipulation,'' \emph{Communications of the ACM}, vol.~16, no.~6, pp.
  372--378, 1973.

\bibitem{mezard2009information}
M.~Mezard and A.~Montanari, \emph{Information, physics, and computation}.\hskip
  1em plus 0.5em minus 0.4em\relax Oxford University Press, 2009.

\bibitem{javanmard2016phase}
A.~Javanmard, A.~Montanari, and F.~Ricci-Tersenghi, ``Phase transitions in
  semidefinite relaxations,'' \emph{Proceedings of the National Academy of
  Sciences}, vol. 113, no.~16, pp. E2218--E2223, 2016.

\bibitem{turk1989interactive}
G.~Turk, ``Interactive collision detection for molecular graphics,'' Ph.D.
  dissertation, The University of North Carolina at Chapel Hill, 1989.

\bibitem{boykov2004experimental}
Y.~Boykov and V.~Kolmogorov, ``An experimental comparison of min-cut/max-flow
  algorithms for energy minimization in vision,'' \emph{IEEE transactions on
  pattern analysis and machine intelligence}, vol.~26, no.~9, pp. 1124--1137,
  2004.

\bibitem{bernardin2008evaluating}
K.~Bernardin and R.~Stiefelhagen, ``Evaluating multiple object tracking
  performance: the clear mot metrics,'' \emph{EURASIP Journal on Image and
  Video Processing}, vol. 2008, no.~1, pp. 1--10, 2008.

\bibitem{opencv_library}
G.~Bradski, ``Opencv library,'' \emph{Dr. Dobb's Journal of Software Tools},
  2000.

\bibitem{sobral2014comprehensive}
A.~Sobral and A.~Vacavant, ``A comprehensive review of background subtraction
  algorithms evaluated with synthetic and real videos,'' \emph{Computer Vision
  and Image Understanding}, vol. 122, pp. 4--21, 2014.

\bibitem{hartley2003multiple}
R.~Hartley and A.~Zisserman, \emph{Multiple view geometry in computer
  vision}.\hskip 1em plus 0.5em minus 0.4em\relax Cambridge university press,
  2003.

\bibitem{munkres1957algorithms}
J.~Munkres, ``Algorithms for the assignment and transportation problems,''
  \emph{Journal of the society for industrial and applied mathematics}, vol.~5,
  no.~1, pp. 32--38, 1957.

\end{thebibliography}

\vspace{\fill}

\end{document}